\documentclass[letterpaper, 10 pt, conference]{ieeeconf}  

\IEEEoverridecommandlockouts                              

\usepackage[usenames]{color}
\usepackage[svgnames]{xcolor}
\definecolor{DarkGreen}{rgb}{0,0.5,0}
\definecolor{DarkRed}{rgb}{0.75,0,0}

\usepackage{amsmath} 
\usepackage{nameref}
\usepackage{amssymb}
\usepackage{graphicx}
\usepackage{tikz}
\usetikzlibrary{shapes,snakes}
\usepackage{comment}
\usepackage{url}
\usepackage[ruled,vlined,linesnumbered]{algorithm2e}
\usepackage{balance}
\usepackage[font=small]{caption}

\usepackage{array}

\newcommand{\bee}{\begin{enumerate} \itemsep -1pt \topsep -2pt}
\newcommand{\eee}{\end{enumerate}}

\usepackage[flushmargin]{footmisc}

\usepackage[colorlinks]{hyperref}
\hypersetup{citecolor=Black}
\hypersetup{linkcolor=Black}
\hypersetup{urlcolor=DarkBlue}
\usepackage{caption}

\usepackage{mathtools}
\usepackage[capitalize]{cleveref}
\crefformat{equation}{(#2#1#3)}
\Crefformat{equation}{Equation~(#2#1#3)}
\Crefname{equation}{Equation}{Equations}

\usepackage{bm}
\newcommand{\bb}[1]{\boldsymbol{{#1}}}
\newcommand{\vect}[1]{\boldsymbol{{#1}}}

\newcommand{\dvect}[1]{\boldsymbol{{\dot{#1}}}}
\newcommand{\ddvect}[1]{\boldsymbol{{\ddot{#1}}}}
\newcommand{\dddvect}[1]{\boldsymbol{{\dddot{#1}}}}

\definecolor{MydarkRed}{RGB}{183,21,33}
\definecolor{MydarkGreen}{RGB}{59,143,50}
\definecolor{Myred}{RGB}{255,0,0}
\definecolor{MyorangeDarker}{RGB}{255,127,42}
\definecolor{MygreenDark}{RGB}{55,200,55}
\definecolor{MyorangeDark}{RGB}{255,212,42}
\definecolor{MyblueLight}{RGB}{85,221,255}
\definecolor{Myblue}{RGB}{0,0,255}
\newcommand{\tikzrectangle}[2][black,fill=red]{\tikz[baseline=0.0ex, line width=0.2mm]\draw[#1] [#1] (0,0) rectangle (0.2,0.2);}%

\newcommand{\subparagraph}{}
\usepackage{titlesec}
\titlespacing{\section}{8pt}{7pt}{6pt}

\renewcommand{\baselinestretch}{.98}
\let\emptyset\varnothing

\title{\LARGE \bf
Real-Time Planning with Multi-Fidelity Models for Agile Flights in Unknown Environments
}

\author{Jesus Tordesillas$^{1}$, Brett T. Lopez$^{1}$, John Carter$^{2}$, John Ware$^{2}$ and Jonathan P. How$^{1}$
	\thanks{$^{1}$J.~Tordesillas, B.~Lopez, J.~How are with the Aerospace Controls Laboratory, MIT, 77 Massachusetts Ave., Cambridge, MA, USA \tt\{jtorde, btlopez, jhow\}@mit.edu}
	\thanks{$^{2}$ J.~Carter and J.~Ware are with the MIT Robust Robotics Group. \tt\{jakeware, jcarter\}@csail.mit.edu}
}
\begin{document}

\maketitle
\thispagestyle{empty}
\pagestyle{empty}

\begin{abstract}
Autonomous navigation through unknown environments is a challenging task that entails real-time localization, perception, planning, and control. UAV’s with this capability have begun to emerge in the literature with advances in lightweight sensing and computing. Although the planning methodologies vary from platform to platform, many algorithms adopt a hierarchical planning architecture where a slow, low-fidelity global planner guides a fast, high-fidelity local planner. However, in unknown environments, this approach can lead to erratic or unstable behavior due to the interaction between the global planner, whose solution is changing constantly, and the local planner; a consequence of not capturing higher-order dynamics in the global plan. This work proposes a planning framework in which multi-fidelity models are used to reduce the discrepancy between the local and global planner. Our approach uses high-, medium-, and low-fidelity models to compose a path that captures higher-order dynamics while remaining computationally tractable. In addition, we address the interaction between a fast planner and a slower mapper by considering the sensor data not yet fused into the map during the collision check. This novel mapping and planning framework for agile flights is validated in simulation and hardware experiments, showing replanning times of 5-40 ms in cluttered environments.
\end{abstract}

\section{INTRODUCTION}

UAV autonomous navigation in unknown environments has received special interest in the last few years because of its unlimited applications, ranging from aerial surveying and inspection to search and rescue. However, these applications are often reduced to low-speed flights due to the current limitations and low rates of the state-of-the-art mappers and planners. The inherent non-convexity of the path planning optimization problem, together with the high mapping and planning rate needed for agile flights make this problem especially hard. This work presents a novel framework to perform high-rate mapping and planning in unknown environments suitable for agile maneuvers, addressing the fundamental problem between the interaction of a global planner and a local planner.

Computational tractability of the planning problem leads to the use of a low-fidelity global planner that computes a cost-to-go (CTG) needed by the high-fidelity local planner. However, the fact that the global planner does not account for the dynamics results in erratic behaviors when the world model is changing rapidly. There is therefore a need of an accurate CTG calculation that captures both the global environment and the dynamic feasibility, maintaining relatively low computation times at the same time. 

\begin{figure}[t]
	\centering
	\includegraphics[width=\columnwidth]{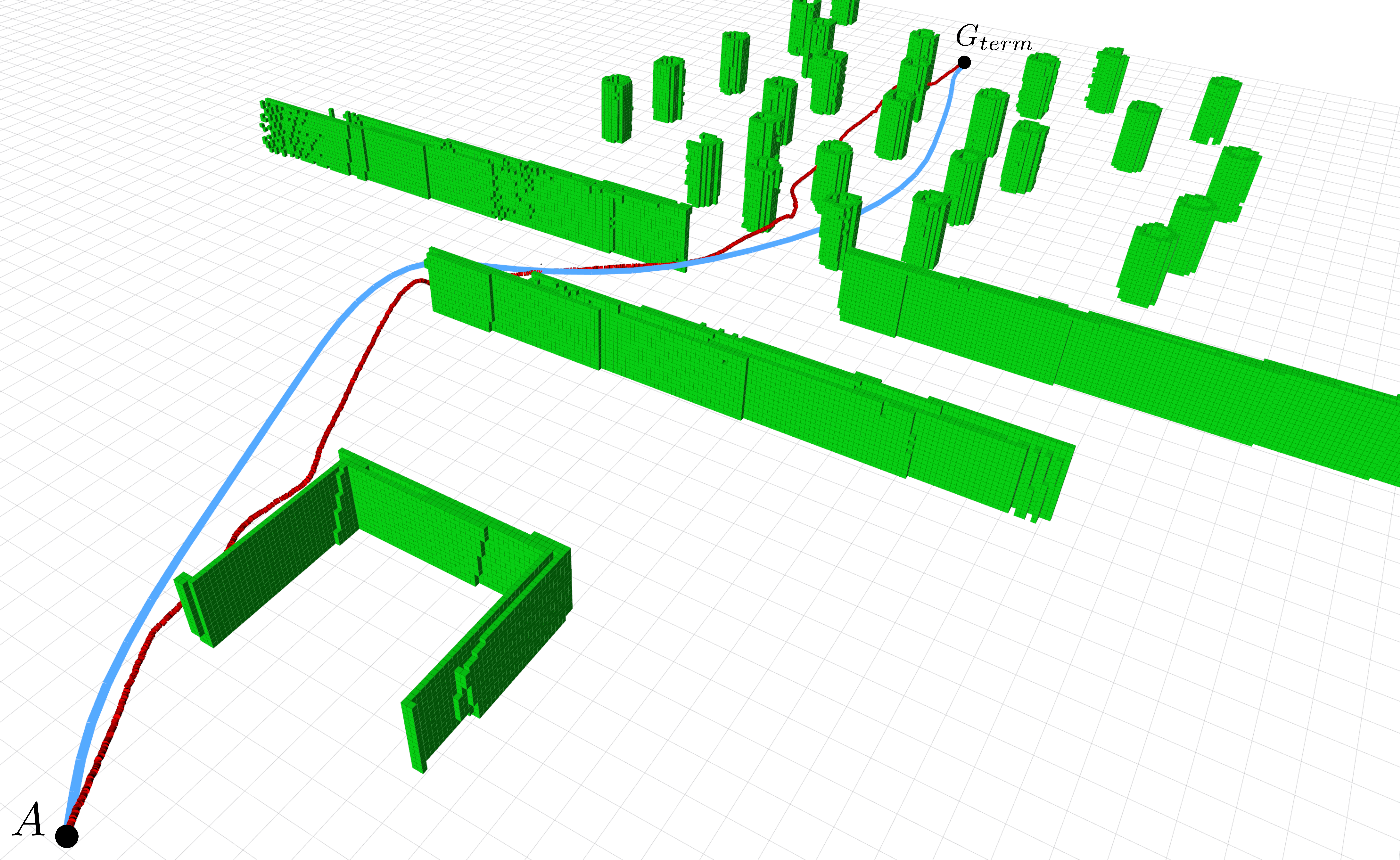}
	\caption{Global optimum and our method. When the map is completely known, the optimal trajectory computed using the approach of \cite{liu2018searchBasedSE3} is shown in blue (\tikzrectangle[black,fill=MyblueLight]{10pt}). The red trajectory (\tikzrectangle[black,fill=Myred]{10pt}) is the solution found by our method, where the world is not known and it is being discovered as the UAV flies forward. The grid is $1$m $\times 1$m, and the sensing range is 10 m.}
	\vskip -0.1in
	\label{fig:upenn}
\end{figure} 

Moreover, the choice of the representation of the environment and the size of the ``global'' map (larger scale than the sensor FOV and the local representation, but typically does not contain all information observed to reduce effort) have a significant impact on the computational cost, but for most systems updates of these models cannot be done at the sensor frame rates ($\sim \!\! 30$ Hz) and updates are typically slower than the re-plan rate. Thus a second design challenge is how to combine the global knowledge (available at a slower rate) with the high-rate local information in the planner representation of the environment. Finally, the state-of-the-art mappers and planners run onboard at $\sim\!\! 5$ Hz, so the the third key challenge is how to optimize the planning and mapping algorithms to achieve higher rates, suitable for aggressive flights.

This work addresses these challenges with the following contributions:
\begin{itemize}
	\item A novel formulation of the planning problem that takes into account the dynamics of the vehicle in the cost-to-go calculation to solve the negative interaction that usually occurs between the global and local planners when operating in unknown environments.
	\item A lightweight fused-based mapping framework using a sliding map to reduce the estimation error influence that runs onboard fusing a depth image in 50 ms.
	\item An integration of a high-rate planner with a slower-rate mapper, with a collision check algorithm that accounts for both the most recent fused information and the available sensed data not included in that map. 
	\item Simulation and hardware experiments showing agile flights in completely unknown cluttered environments, achieving replanning times of 5-40 ms. 

\end{itemize}

\section{RELATED WORK}
\label{sec:related_work}

Different methods have been proposed in the literature for planning, mapping, and the integration of this two.  

For \textbf{planning}, most of the current state-of-the-art methods exploit the differential flatness of the quadrotors, and solve the planning problem minimizing the squared norm of a derivative of the position \cite{mellinger2011minimum},  \cite{van1998real} to find a dynamically feasible smooth trajectory \cite{richter2016polynomial}. On the one hand, there are approaches where the obstacle constraints, and sometimes also input constraints, are checked after solving the optimization problem: Some of them use stitched polynomial trajectories that pass through several waypoints obtained running RRT-based methods \cite{mellinger2011minimum}, \cite{richter2016polynomial}, \cite{loianno2017estimation}. Others use Pontryagin's Minimum Principle to find closed-form solutions \cite{mueller2015computationally}. These closed-form solutions are also used to search over the state space \cite{liu2017searchminimumT}, \cite{liu2018searchBasedSE3}. Alternatively, there are works that use the cost function to penalize the distance to the obstacles \cite{oleynikova2016continuous}, \cite{oleynikova2018safe}, requiring usually computationally expensive distance fields representations. Finally, there are also approaches that add the obstacle constraints in the optimization problem, like Mixed-Integer Programming (MIP) \cite{richards2002aircraft}, convex decompositions \cite{liu2018convex}, \cite{liu2017planning}, \cite{wattersontrajectory} and Successive Convexification \cite{mao2018successive}, \cite{augugliaro2012generation}, \cite{schulman2014motion}. 

Most of the approaches described above either assume that the global map is known, and the optimization is done using the global map. Others assume that there is a global planner (RRT or exploration based planners) that gives the waypoints to the local planner. In agile flights, this leads to oscillatory behaviors, since the world is being discovered and the solution of the global planner changes constantly. 

Moreover, two main categories can be highlighted in the \textbf{mapping} methods proposed in the literature: memory-less and fused-based methods. The first category includes the approaches that rely only on instantaneous sensing data, using only the last measurement, or weighting the data \cite{dey2016vision}, \cite{florence2018nanomap}. These approaches are in general unable to reason about obstacles observed in the past \cite{lopez2017aggressive3D}, \cite{lopez2017aggressivelimitedFOV}. The second category is the fusion-based approach, in which the sensing data are fused into a map, usually in the form of an occupancy grid or distance fields \cite{lau2010improved}, \cite{oleynikova2017voxblox}. Two drawbacks of these approaches are the influence of the estimation error, and the fusion time (which means that the planner usually uses an out-of-date fused map).

Finally, several approaches have been proposed for the \textbf{integration} between the planner and the mapper: reactive and map-based planners. Reactive planners often use a memory-less representation of the environment, and closed-form primitives are usually chosen for planning  \cite{lopez2017aggressive3D} and \cite{lopez2017aggressivelimitedFOV}. These approaches often fail in complex cluttered scenarios. On the other hand, map-based planners usually use occupancy grids or distance fields to represent the environment. These planners either plan all the trajectory at once or implement a Receding Horizon Planning Framework, optimizing trajectories locally and based on a global planner. Moreover, when unknown space is also taken into consideration, several approaches are possible: \cite{pivtoraiko2013incremental} and \cite{chen2016online} used optimistic planners (considering unknown space as free), while in \cite{oleynikova2016continuous} and \cite{oleynikova2018safe}, an optimistic global planner is used combined with a conservative local planner.

\section{PROBLEM FORMULATION}
\label{sec:problem_formulation}

\subsection{Planning}
Computing a dynamically feasible trajectory from the start to the goal is typically intractable, which is the standard argument for the distinction between the global and local planners. The global planner gives the local planner a notion of cost-to-go (CTG) or traversability in certain directions; while the local planner deviates around the nearby obstacles and chooses the terminal point accordingly. Key to this CTG calculation is the trade-off between accuracy and computation time. Issues include the sophistication of the dynamics used in the calculation and the number of points at which the CTG is computed. 

In our proposed framework, Jump Point Search (JPS)  is used as a global planner to find the shortest path from the current position to the goal. JPS was chosen instead of A* because it runs an order of magnitude faster, while still guaranteeing completeness and optimality \cite{harabor2011online}, \cite{liu2017planning}. The only assumption of JPS is a uniform grid, which holds in our case. JPS is only done for position (not velocity or acceleration) to reduce the computational burden. 

Our local planner monitors the JPS solution for drastic changes between each replan. If a large change is detected, a new path, composed of a high, medium, and low fidelity model (where the model order is reduced farther away from the vehicle) is created for the current and last JPS solution. This procedure is able to capture a subset of the dynamics while maintaining computational tractability. The resulting multi-fidelity paths are compared and the path with lowest cost is selected for execution. This hierarchical trajectory consists of a jerk-controlled part, a velocity-controlled part, and a geometric part. 

The jerk-controlled primitive is the part of the trajectory nearest to the current position, and it is the one that will be actually executed. The quadrotor is modeled using triple integrator dynamics with state
$
\mathbf{x}^T = \left[ \vect{x}^T ~ \dvect{x}^T ~ \ddvect{x}^T ~ \right] = \left[\vect{x}^T ~\vect{v}^T ~\vect{a}^T\right]
$
and control input $\mathbf{u} = \dddvect{x} = \vect{j}$ (where $\vect{x}$, $\vect{v}$, $\vect{a}$, and $\vect{j}$ are the vehicle's position, velocity, acceleration, and jerk, respectively), and the following convex optimization problem is solved using CVXGEN \cite{mattingley2012cvxgen} to find this trajectory:
\begin{align}
\min_{u_{0:N-1}} & \sum_{i=0}^{N-1}\left\Vert \mathbf{u}_{i}\right\Vert ^{2}+(\mathbf{x}_{N}-\mathbf{x}_{f})^{T}Q(\mathbf{x}_{N}-\mathbf{x}_{f}) \label{eq:1} \\
\textrm{subject to } & \mathbf{x}_{0}=\mathbf{x}_{init}\nonumber\\
& \mathbf{x}_{k+1}=M_{1}\mathbf{x}_{k}+M_{2}\mathbf{u}_{k}  \quad \forall k=0:N-1 & \nonumber\\
& \left\Vert \boldsymbol{v}_{k}\right\Vert _{\infty}\le v_{max} \qquad \quad \; \forall k=1:N &\nonumber\\
& \left\Vert \boldsymbol{a}_{k}\right\Vert _{\infty}\le a_{max} \qquad \quad \; \forall k=1:N &\nonumber\\
& \left\Vert \boldsymbol{u}_{k}\right\Vert _{\infty}\le j_{max} \qquad \quad \; \forall k=0:N-1. & \nonumber
\end{align}
In this problem, the number of discretization steps $N$ is fixed. The time step $dt$ (embedded in $M_{1}$ and $M_{2}$) is computed as
$$dt=\max\{T_{v_x},T_{v_y},T_{v_z}, T_{a_x},T_{a_y},T_{a_z},T_{j_x},T_{j_y},T_{j_z}\}/N$$
where $T_{v_i}$, $T_{a_i}$ and $T_{j_i}$ are the solution for each axis $i=\{x,y,z\}$ of the constant-velocity, constant-acceleration and constant-jerk motion equations applying $v_{max}$, $a_{max}$ and $j_{max}$ respectively. This $dt$ is a tight lower bound that is increased in each iteration until the problem converges.

The second key part of the trajectory is a velocity-controlled primitive. It is a trade-off between the sophistication of the dynamic model (lower order than jerk input model), but sufficiently accurate to capture the UAV CTG (more accurate than the distance-based cost). Moreover, it ensures that the computation times are maintained in the order of 300 $\mu$s, four times faster than the ones required when the input is higher. The final part of the trajectory is the part of the JPS solution that goes from the end of the velocity-controlled primitive to the goal. This provides an indication of how to avoid traps and avoid obstacles, but there is little attempt to capture to vehicle dynamics at that distance away -- that is done when the receding horizon controller gets closer.

\subsection{Mapping}
\begin{figure}[t]
	\centering
	\includegraphics[width=\columnwidth]{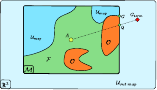}
	\caption{Sliding map and goal projection. The sliding map $\mathcal{M}$ has occupied space $\mathcal{O}$, unknown space $\mathcal{U}_{map}$ and free space $\mathcal{F}$. Total unknown space is $\mathcal{U}=\mathcal{U}_{map}\cup\mathcal{U}_{out\;map}$. $\bb{Q}$ is the projection into the map of the terminal goal $\bb{G}_{term}$ in the direction of $\protect\overrightarrow{\bb{AG}_{term}}$. The closest free or unknown frontier point to $\bb{Q}$ ($\bb{G}$ in the figure) is selected as the goal. }
	\vskip -0.1in
	\label{fig:space}
\end{figure}

A sliding map, which moves with the UAV, is used to represent the world. This is a compromise between storing the whole world and relying only on local maps. It also tries to minimize the accumulated estimation error. This map $\mathcal{M}$ contains free space $\mathcal{F}$, (known) obstacles $\mathcal{O}$ and unknown space $\mathcal{U}_{map}$ (see Fig.~\ref{fig:space}). In this way, $$\mathbb{R}^3=\mathcal{O}\cup\mathcal{F}\cup\mathcal{U}_{map}\cup\mathcal{U}_{Out\;map}=\mathcal{M}\cup\mathcal{U}_{Out\;map}$$

Using this map, the collision check for each of the three primitives presented above is done as follows: the jerk-controlled trajectory is considered collision-free if it does not intersect $\mathcal{O}\cup \mathcal{U}$. The velocity-controlled primitive is forced to pass through the waypoints of the JPS solution. Finally, the JPS path is guaranteed not to hit $\mathcal{O}$. In this way, and similar to \cite{oleynikova2018safe}, JPS is an optimistic global planner, while the local planner is conservative.

\section{ALGORITHM}
\label{sec:algorithm_section}

\begin{figure}[t]
	\centering
	\includegraphics[width=\columnwidth]{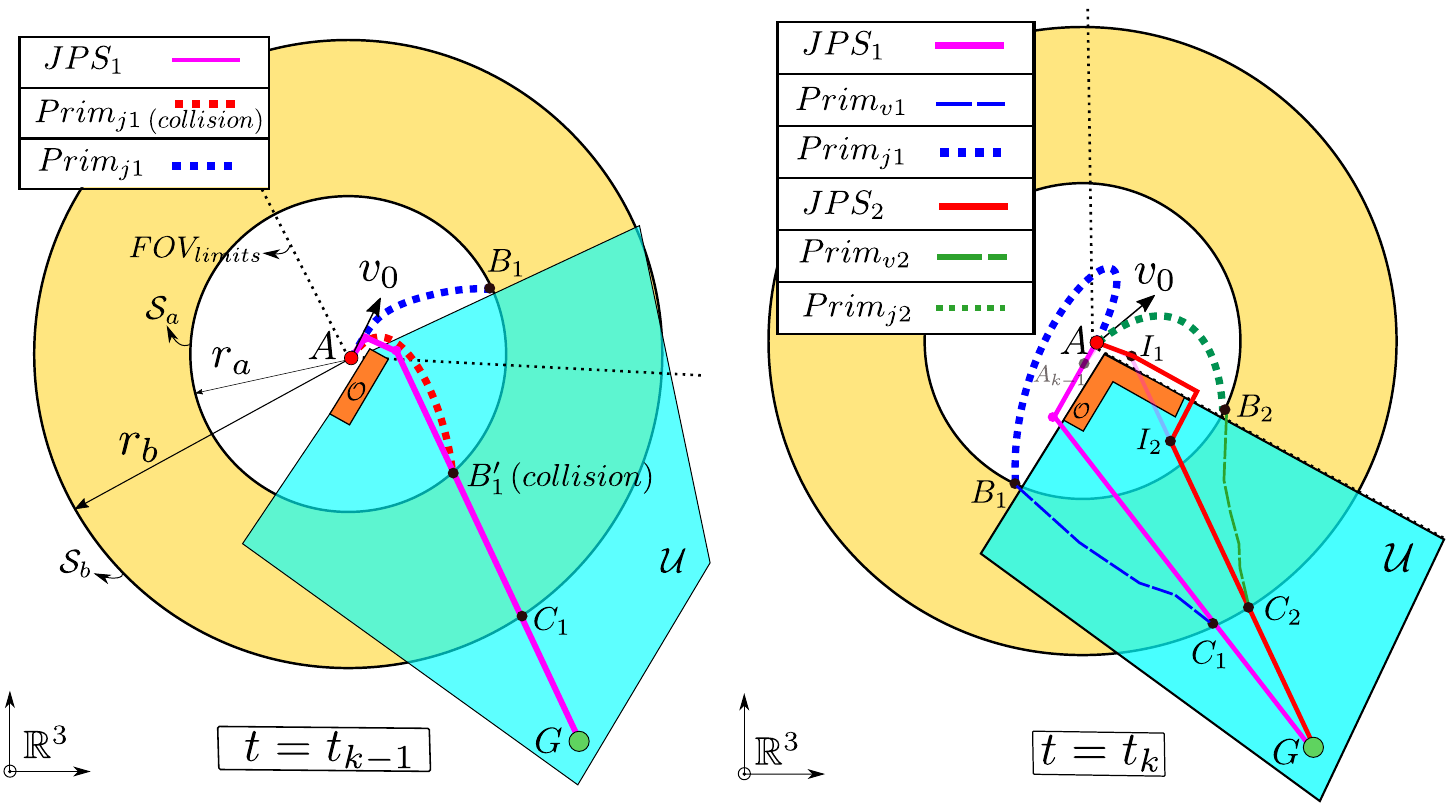}
	\caption{Illustration of the Alg.~\ref{algo: myalgorithm}. The radii $r_{a}$ and $r_{b}$ define the spherical surfaces $\mathcal{S}_{a}$ and $\mathcal{S}_b$. JPS is used to compute the paths $JPS_{1}$ and $JPS_{2}$. The UAV chooses the jerk-controlled primitive $Prim_{ji}$ that has lowest cost, considering the cost of that primitive, the velocity-controlled primitive $Prim_{vi}$ and the distance from $\bb{C}_{i}$ to $\bb{G}$ following $JPS_{i}$. Unknown space $\mathcal{U}$ is shown in blue. Note that the figure is in 2D for visualization purposes, but the planning is in 3D.}
	\vskip -0.1in
	\label{fig:planning_strategy}
\end{figure}

\begin{figure}[t]
	\centering
	\includegraphics[width=\columnwidth]{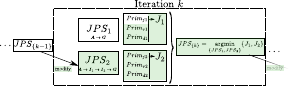}
	\caption{Iteration $k$ in the Alg.~\ref{algo: myalgorithm}. $JPS_{2}$ is the modified version of  $JPS_{\{k-1\}}$, that avoids the new obstacles detected. The algorithm chooses the jerk-controlled primitive $Prim_{ji}$ that has the lowest associated cost-to-go $J_{i}$. The terms in green are only computed if $\angle \bb{B'}_{1}\bb{A}\bb{B'}_{\{k-1\}}$ is greater than $\alpha_{0}$. }
	\vskip -0.1in
	\label{fig:iteration}
\end{figure}


Let us introduce some notation first (see also Fig. \ref{fig:planning_strategy}): Let $\bb{A}$ be the state taken (in the previous planned trajectory) some steps ahead of the current position of the UAV. Let $\mathcal{S}_a$ and  $\mathcal{S}_b$ be two concentric spheres with center in $\bb{A}$, and with radius $r_{a}$ and $r_{b}$ respectively. $JPS$ will denote the shortest piece-wise linear path found by running JPS between $\bb{A}$ and the goal $\bb{G}$. For this path, define the tuples
$JPS_{wp}\coloneqq(\bb{q}_{1},\bb{q}_{2},\ldots,\bb{q}_{n})$ and 
$JPS_{path}\coloneqq (\overline{\bb{q}_{1}\bb{q}_{2}},\ldots,\overline{\bb{q}_{n-1}\bb{q}_{n}})$,
where $JPS_{wp}$ has the $n$ waypoints $\boldsymbol{q}_{i}\in\mathbb{R}^{3}$  of the solution of JPS  (being $\bb{q}_{1}$ the start and $\bb{q}_{n}$ the goal) and $JPS_{path}$ contains the segments between these waypoints. 

Define the point $\bb{B}'\coloneqq\cap_{\bb{f}}(\mathcal{S}_{a}, JPS_{path})$, where $\cap_{\bb{f}}(\mathcal{S}, JPS_{path})$ is a function that computes the \textbf{f}irst intersection between the spherical surface $\mathcal{S}$ and the path $JPS_{path}$. Similarly, $\bb{C} \coloneqq \cap_{\bb{l}}(\mathcal{S}_{b}, JPS_{path})$ will be the $\boldsymbol{l}$ast intersection. In an analogous way, $\cap_{\boldsymbol{int}}(\mathcal{S}_{a},\mathcal{S}_{b}, JPS_{wp}) $ 
will denote all the elements of  $JPS_{wp}$ that are $\boldsymbol{int}$ermediate between $\bb{B'}$ and $\bb{C}$. 
Hence, $JPS_{wp}$ can be written as: $$JPS_{wp}=(\bb{q}_{1},...\bb{q}_{r-1},\underbrace{\bb{q}_{r},\ldots,\bb{q}_{s-1}}_{\cap_{\boldsymbol{int}}(\mathcal{S}_{a},\mathcal{S}_{b}, JPS_{wp})},\bb{q}_{s},\ldots,\bb{q}_{n})$$ where $\bb{q}_{r}$ is the first point outside the sphere $\mathcal{S}_{a}$ and $\bb{q}_{s-1}$ is the last point inside the sphere $\mathcal{S}_{b}$. The subindex $\{\cdot\}$ will indicate the iteration number. Finally, the subindex 1 in $JPS$ will denote the JPS solution of the current iteration, while 2 will be the modified (i.e not intersecting any obstacles) version of $JPS_{\{k-1\}}$ (chosen in the previous iteration). Same applies to $\bb{B'}$ and $\bb{C}$. The concept of modified is explained later on. 

In our algorithm, the whole trajectory planned is divided into three primitives: $Prim_{j}\cup Prim_{v}\cup Prim_{d}$, which are defined as follows: 
\bee
\item $Prim_{j}$ is a jerk-controlled primitive that has $\bb{A}$ as initial state and a final stop condition in the point $\bb{B}$, which is defined  as the point in the sphere $\mathcal{S}_a$ near to $\bb{B}'$ that guarantees that $Prim_{j} \cap (\mathcal{O}\cup\mathcal{U})=\emptyset$.
\item $Prim_{v}$ is defined as the composition of several velocity-controlled primitives between each pair of consecutive points of $(\bb{B},\bb{q}_{r},\ldots,\bb{q}_{s-1},\bb{C})$. 
\item Finally, $Prim_{d}$ is the part of $JPS$ that goes from $\bb{C}$ to the goal $\bb{G}$, and therefore $Prim_{d}\cap\mathcal{O}= \emptyset$. 
\eee
The time-normalized costs associated with each one of these primitives are:
\begin{equation*}
\begin{aligned}\underset{}{} & J_{Prim_{j}}=\frac{N\cdot dt}{j_{max}^{2}}\cdot\sum\limits_{i=0}^{N-1}\left\Vert \boldsymbol{j}_{i}\right\Vert _{2}^{2}\\
& J_{Prim_{v}}=\sum\limits_{k=0}^{s-r} \left(\frac{N\cdot dt_{k}}{v_{max}^{2}} \cdot\sum\limits_{i=0}^{N-1}\left\Vert \boldsymbol{v}_{k_{i}}\right\Vert _{2}^{2}\right)\\
& J_{Prim_{d}}=\frac{\left\Vert \boldsymbol{q}_{s}-\boldsymbol{C}\right\Vert _{2}}{v_{max}}+\sum\limits_{k=s}^{n-1}\left(\frac{\left\Vert \boldsymbol{q}_{k+1}-\boldsymbol{q}_{k}\right\Vert _{2}}{v_{max}}\right)
\end{aligned}
\end{equation*}
The total cost of the planned trajectory is then $J\coloneqq J_{Prim_{j}}+J_{Prim_{v}}+J_{Prim_{d}}$. In this total cost we combine a high-fidelity model near the UAV, a medium-fidelity model in the medium part, and model without dynamics for the farthest part of the trajectory. This cost is an approximation of the total cost of the jerk-controlled trajectory that goes from $\bb{A}$ to $\bb{G}$, but it is much faster to compute. Also, it is more accurate than relying only on jerk for the first part, and distance in the JPS path for the rest, since an intermediate velocity-controlled primitive is included.

\begin{algorithm}[t]
	\footnotesize
	
	\DontPrintSemicolon
	\KwData{$A$, $G_{term}$, $\mathcal{O}$, $\mathcal{F}$, $\mathcal{U}$, $\alpha_{0}>0$, $R_{b}>R_{a,max}>R_{a,min}>0$   }
	
	\SetKwFunction{FMain}{\textcolor{ForestGreen}{\textbf{Replan}}}
	\SetKwProg{Pn}{Function}{:}{\KwRet}
	\Pn{\FMain{}}{
		
		$k\leftarrow k+1 $ , $J_{1}\leftarrow 0 \text{~and~}  J_{2}\leftarrow\infty$ \;
		$G\leftarrow $ Project Terminal Goal $G_{term}$  \label{projection}\;
		$JPS_{1} \leftarrow$ Run JPS $A\rightarrow G$\;
		$r_{a} \leftarrow \min(saturate_{R_{a,min}}^{R_{a,max}}(\| \overrightarrow{Aq_{2}} \|_{2}),  \| \overrightarrow{AG} \|_{2})$\;
		$r_{b} \leftarrow \min(R_b,   \| \overrightarrow{AG} \|_{2})$\;

		$Prim_{j1} \leftarrow \textcolor{blue}{GetPrimj}(1)$\;
	
					\If{$\angle B'_{1}AB'_{\{k-1\}}> \alpha_{0}$ }{

		$JPS_{2}\leftarrow JPS_{\{k-1\}}$ \;
		\If{$JPS_{\{k-1\}}\cap \mathcal{O}\neq \emptyset$ \label{intersectionbegin}}{\ 
			$JPS_{2} \leftarrow$ Run JPS $A\rightarrow I_{1}\rightarrow I_{2} \rightarrow G$ \label{intersectionend} 
		}

   $Prim_{j2} \leftarrow \textcolor{blue}{GetPrimj}(2)$\;  \label{getPrimj2}
	$J_{1}\leftarrow \textcolor{red}{GetCost}(1) \text{~and~} J_{2}\leftarrow \textcolor{red}{GetCost}(2)$\;

	}
		Choose $i$ with lowest cost $J_{i}$\;  \label{startdecision}
		$JPS_{\{k\}}\leftarrow JPS_{i}$  \text{~and~}  	$B'_{\{k\}}\leftarrow B'_{i}$  \;
     	\Return $Prim_{ji}$   \label{enddecision}

	}
\footnotesize
	\SetKwFunction{FMain}{\textcolor{blue}{\textbf{GetPrimj}}}
\SetKwProg{Fn}{Function}{:}{}
\Fn{\FMain{$i$}}{

	$B'_{i}\leftarrow \cap_{f}(\mathcal{S}_{a}, JPS_{i,path})$\;	
	$\mathcal{K}\leftarrow SamplePoints(B'_{i})$\label{sample_priority_queue}\;
	\For{$Size(\mathcal{K})$ times\label{startloop1}}{ 
		$B_{i} \leftarrow Pop(\mathcal{K})$\;
		$ Prim_{ji}	\leftarrow$ Compute Jerk Primitive $A \rightarrow B_{i}$\;
		\textbf{return} $Prim_{ji}$ \textbf{if} $Prim_{ji} \cap (\mathcal{O}\cup \mathcal{U})=\emptyset$  \label{endloop1}
	}
}
\SetKwFunction{FMain}{\textcolor{red}{\textbf{GetCost}}}
\SetKwProg{Pn}{Function}{:}{\KwRet}
\Pn{\FMain{$i$}}{
	
	$C_{i}\leftarrow \cap_{l}(\mathcal{S}_{b}, JPS_{i,path})$\;
	$\mathcal{WP} \leftarrow \cap_{int}(\mathcal{S}_{a},\mathcal{S}_{b}, JPS_{i,wp})$\;
	$ Prim_{vi}\leftarrow$ Compute Vel. Primitives $B_{i}\rightarrow \mathcal{WP}\rightarrow C_{i}$\;  
	\Return$ J_{ Prim_{j_{i}}}+J_{ Prim_{v_{i}}}+J_{ Prim_{d_{i}}}$\;
	
}
	\normalsize
	\caption{Replan \label{IR}}
	\label{algo: myalgorithm}
\end{algorithm}

\begin{figure}[t]
	\centering
	\includegraphics[width=7cm]{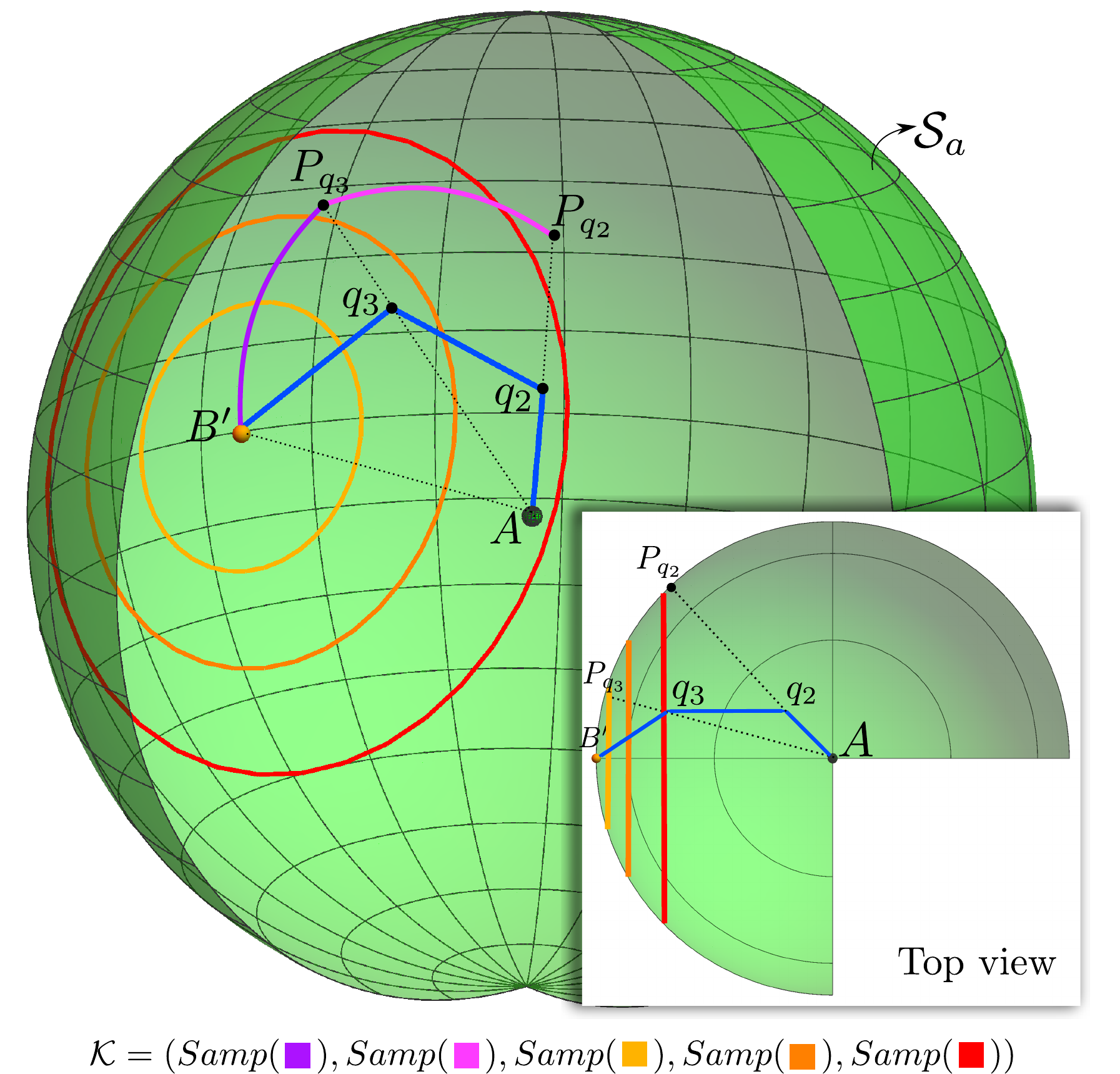}
	\caption{Priority queue $\mathcal{K}$. Given $JPS$ (\tikzrectangle[black,fill=Myblue]{10pt}), the priority queue $\mathcal{K}$ returned by $SamplePoints(\bb{B'})$ contains points in this order: First samples along the spherical arc $\bb{B'} \rightarrow \bb{P}_{\bb{q}_{3}}$. Then along the arc from $\bb{P}_{\bb{q}_{3}}$ to the next projection, and so on. After that, several samples are taken from concentric circumferences to $\bb{B'}$. }
	\vskip -0.1in
	\label{fig: priorityqueue}
\end{figure}

The proposed approach is shown in Alg.~\ref{algo: myalgorithm} and Fig. \ref{fig:planning_strategy}. For iteration $k$, it proceeds as follows --- first (line \ref{projection}), the terminal goal $\bb{G_{term}}$ is projected into $\mathcal{M}$ in the direction of $\overrightarrow{\bb{AG_{term}}}$ to obtain $\bb{Q}$ (see Fig.~\ref{fig:space}). The nearest unknown or free point to $\bb{Q}$ is selected as the (intermediate) goal $\bb{G}$, and JPS is run from the actual position $\bb{A}$ to $\bb{G}$ to obtain $JPS_{1}$. The intersection of $JPS_{1}$ with the inner sphere $\mathcal{S}_{a}$ defines the point $\bb{B'}_{1}$. This point indicates the direction towards which the jerk-controlled planner should optimize. As a jerk-controlled primitive from $\bb{A}$ to $\bb{B'}_{1}$ is not guaranteed to be collision free, we sample $\approx 30$ points in the sphere $\mathcal{S}_{a}$ around $\bb{B}'_{1}$ to obtain a final position that makes this primitive collision-free, storing them in the priority queue $\mathcal{K}$ (line \ref{sample_priority_queue}). We prioritize points near to $\bb{B'}$ that have at the same time a high probability of the primitive being collision-free. As the geometry of $JPS$ encodes where the obstacles are, we sample these points in the following way (Fig.~\ref{fig: priorityqueue}): First, the points $\bb{q}_{2}$,\ldots,$\bb{q}_{r-1}$ are projected onto $\mathcal{S}_{a}$ to obtain $\bb{P}_{\bb{q}_{2}}$,\ldots,$\bb{P}_{\bb{q}_{r-1}}$. Then we sample from $\bb{B'}$ to $\bb{P}_{\bb{q}_{r-1}}$, from $\bb{P}_{\bb{q}_{r-1}}$ to $\bb{P}_{\bb{q}_{r-2}}$, and so on. Finally, we append to this priority queue some samples taken in concentric circles to $\bb{B'}$, to increase the probability of finding a feasible final condition. 

Then we iterate over each point in $\mathcal{K}$ (setting it as the final position in the optimization problem \ref{eq:1}) and stop when a collision free primitive (i.e that does not intersect $\mathcal{O}\cup\mathcal{U}$) is found (lines \ref{startloop1}-\ref{endloop1}). At this point, the algorithm computes the angle $\angle \bb{B'}_{1}\bb{A}\bb{B'}_{\{k-1\}}$, where $\bb{B'}_{\{k-1\}}$ is the intersection of $\mathcal{S}_{a}$ with $JPS_{\{k-1\}}$. This angle gives a measure of how much the JPS solution has changed from the iteration $k-1$. A small angle indicates that $JPS_{1}$ and $JPS_{\{k-1\}}$ are very similar (at least within the sphere $\mathcal{S}_a$), and that therefore the local plan will not differ much from the iteration $k-1$. Hence, if this angle is smaller than a threshold $\alpha_{0}$ (typically $\approx 15^{\circ}$), the algorithm finishes, and $Prim_{j1}$ is returned. If it is bigger, the algorithm needs to decide whether obtaining the local plan using $JPS_{1}$, or relying on the previous path $JPS_{\{k-1\}}$. To do this, in lines \ref{intersectionbegin}-\ref{intersectionend} first we modify $JPS_{\{k-1\}}$ by obtaining $\bb{I}_{1}$ and $\bb{I}_{2}$ (first and last intersections of $JPS_{\{k-1\}}$ with $\mathcal{O}$) and run JPS three times to obtain the paths $\bb{A} \rightarrow \bb{I}_{1}$,  $\bb{I}_{1}\rightarrow \bb{I}_{2}$, and $\bb{I}_{2}\rightarrow\bb{G}$. The union of these paths is $JPS_{2}$. Once $JPS_{2}$ is obtained, a very similar process as explained before is done, but with $JPS_{2}$ in this case, to obtain $Prim_{j2}$ (line \ref{getPrimj2}). 

The decision between $Prim_{j1}$ or $Prim_{j2}$ is made (lines \ref{startdecision}-\ref{enddecision}) choosing the one that has the lowest cost $J_{i}\coloneqq J_{ Prim_{ji}}+J_{ Prim_{vi}}+J_{ Prim_{di}}$ , $i=\{1,2\}$. 

\begin{figure}[t]
	\centering
	\includegraphics[width=\columnwidth]{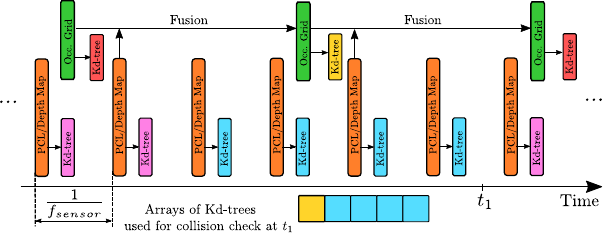}
	\caption{Instantaneous sensing data and occupancy grid pipeline. Sensing data from the depth sensor (\tikzrectangle[black,fill=MyorangeDarker]{10pt}) is received at $f_{sensor}$. New point clouds are fused into the Occupancy Grid (\tikzrectangle[black,fill=MygreenDark]{10pt}). Collision check at $t_{1}$ is done using an array of \textit{k}-d trees that contains the  \textit{k}-d tree of the last map fused (\tikzrectangle[black,fill=MyorangeDark]{10pt}), and the \textit{k}-d trees of some of the last point clouds received that are not included in the map (\tikzrectangle[black,fill=MyblueLight]{10pt}).}
	\vskip -0.1in
	\label{fig:pclouds_pipeline}
\end{figure} 

\subsection{Mapping and Planning Integration}
We fuse a depth map into the occupancy grid using the 3D Bresenham's line algorithm for ray-tracing \cite{bresenham1965algorithm}. However, as discussed earlier, the mapper update rate is slower than the sensor frame rate, and problems can arise when the local planner only relies on this (out-of-date) map to generate a primitive. This issue is addressed here by storing the \textit{k}-d trees of the point clouds that have arrived since the most recent map was published (see Fig. \ref{fig:pclouds_pipeline}). Collision checks are then done using the occupancy grid and some of the saved \textit{k}-d trees of the instantaneous point clouds. This combination ensures that the local planner relies both on the most recent fused map with global knowledge of the world, and on up-to-date point clouds that contain the instantaneous sensing data.

\section{EXPERIMENTAL RESULTS}
\label{sec:experimental_results}

\subsection{Simulation}

We evaluate the performance of the proposed algorithm in different simulated scenarios. The simulator uses C++ custom code for the dynamics engine, and Gazebo \cite{koenig2004design} to simulate perception data (in the form of a depth map and a point cloud). In all these simulations, the depth camera has a horizontal FOV of $90^{\circ}$, and a sensing range of $10$ m. 

\begin{figure}[t]
 	\centering
 	\includegraphics[width=\columnwidth]{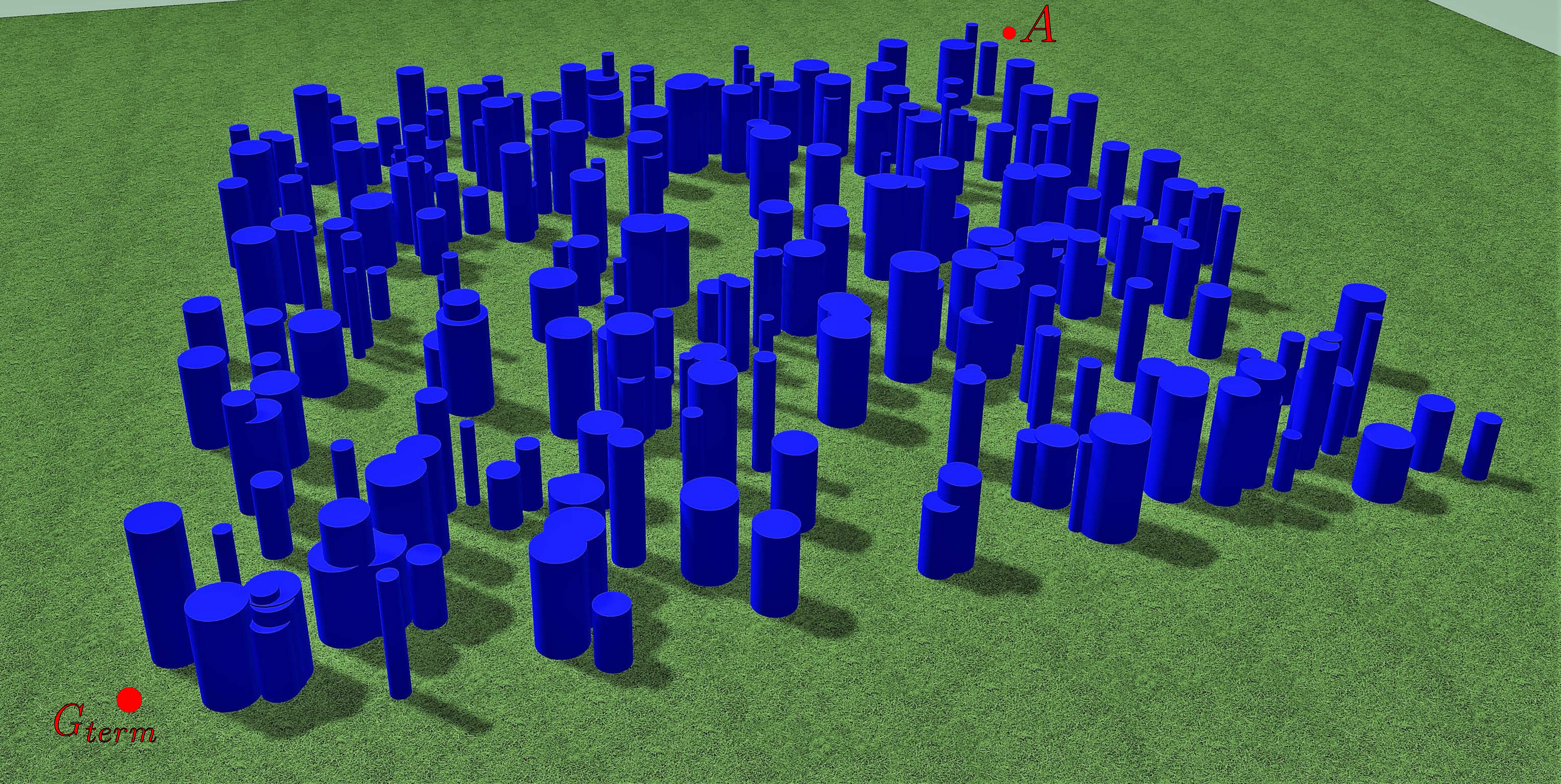}
 	\caption{Forest simulation. The UAV must fly from $\bb{A}$ to $\bb{G_{term}}$ in a $50$m$\times 50$m forest with an obstacle density of $0.1$~obstacles/m$^2$.  }
 	\label{fig:forest}
 %
%
	\centering
		\includegraphics[width=\columnwidth]{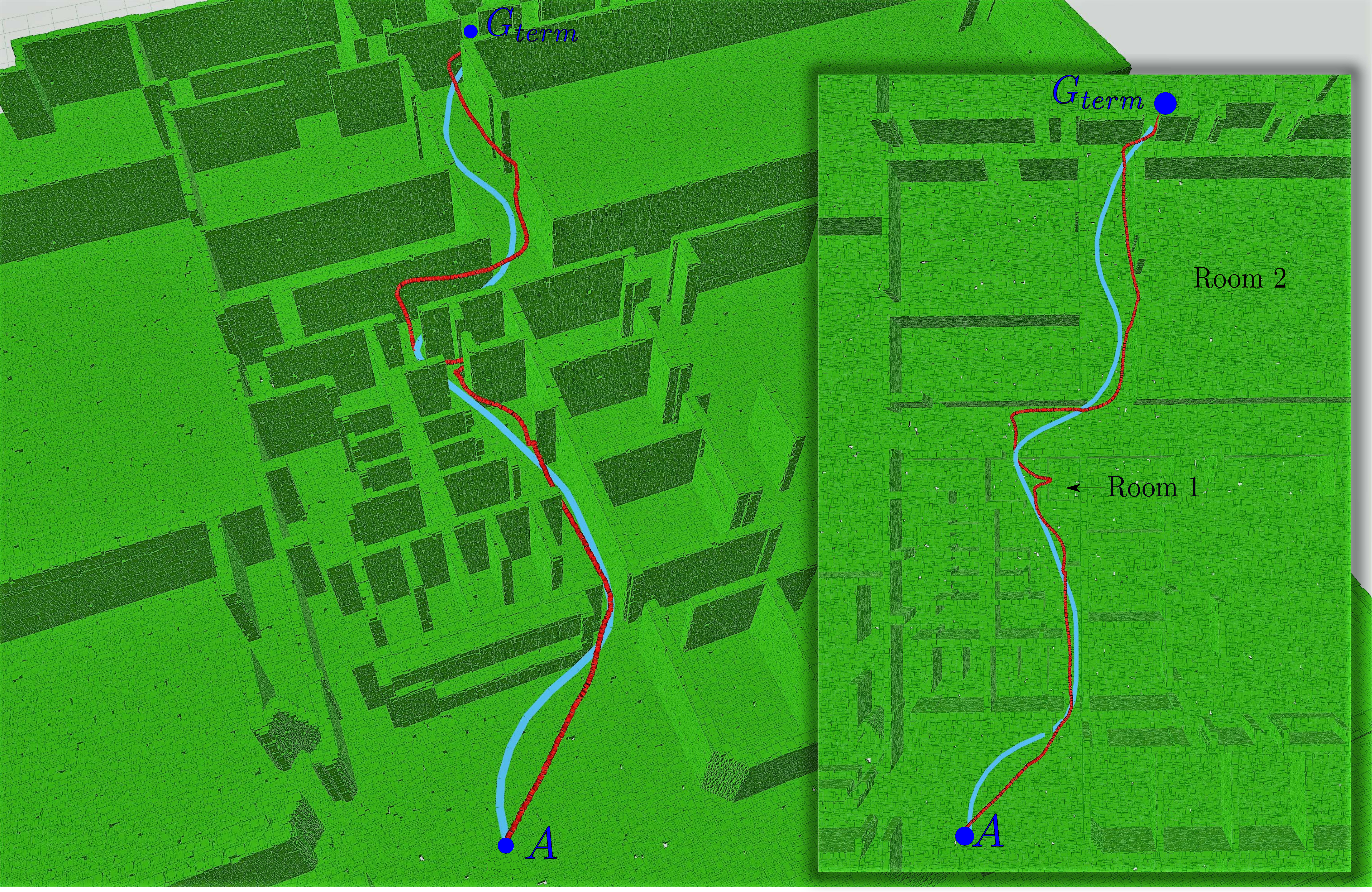}
	\caption{Office simulation. The UAV must fly from $\bb{A}$ to $\bb{G_{term}}$ in an office environment. The optimal trajectory is shown in blue (\tikzrectangle[black,fill=MyblueLight]{10pt}). The red trajectory (\tikzrectangle[black,fill=Myred]{10pt}) is the solution found by our method.}
	\label{fig:office}
\end{figure}

We first compare our method against six different methods: Incremental approach (no goal selection), random goal selection, optimistic RRT$^\star$ (unknown space = free), conservative RRT$^\star$ (unknown space=occupied), ``next-best-view" planner (NBVP) \cite{bircher2016receding}, and Safe Local Exploration \cite{oleynikova2018safe}. These six methods are described deeper in \cite{oleynikova2018safe}. The scenario setting is a random cluttered forest with an obstacle density of $0.1$~obstacles/m$^2$ (see Fig. \ref{fig:forest}) and a sliding map size of $20$m$\times 20$m, with a voxel size of 10 cm. The results for ten different random forests are shown in Table~\ref{tab:table_forest}. Our method succeeds in all 10 simulations and obtains a path that is in average $19$--$47$ \% shorter than the other methods. 

\begin{figure}[t]
	\centering
	\includegraphics[width=\columnwidth]{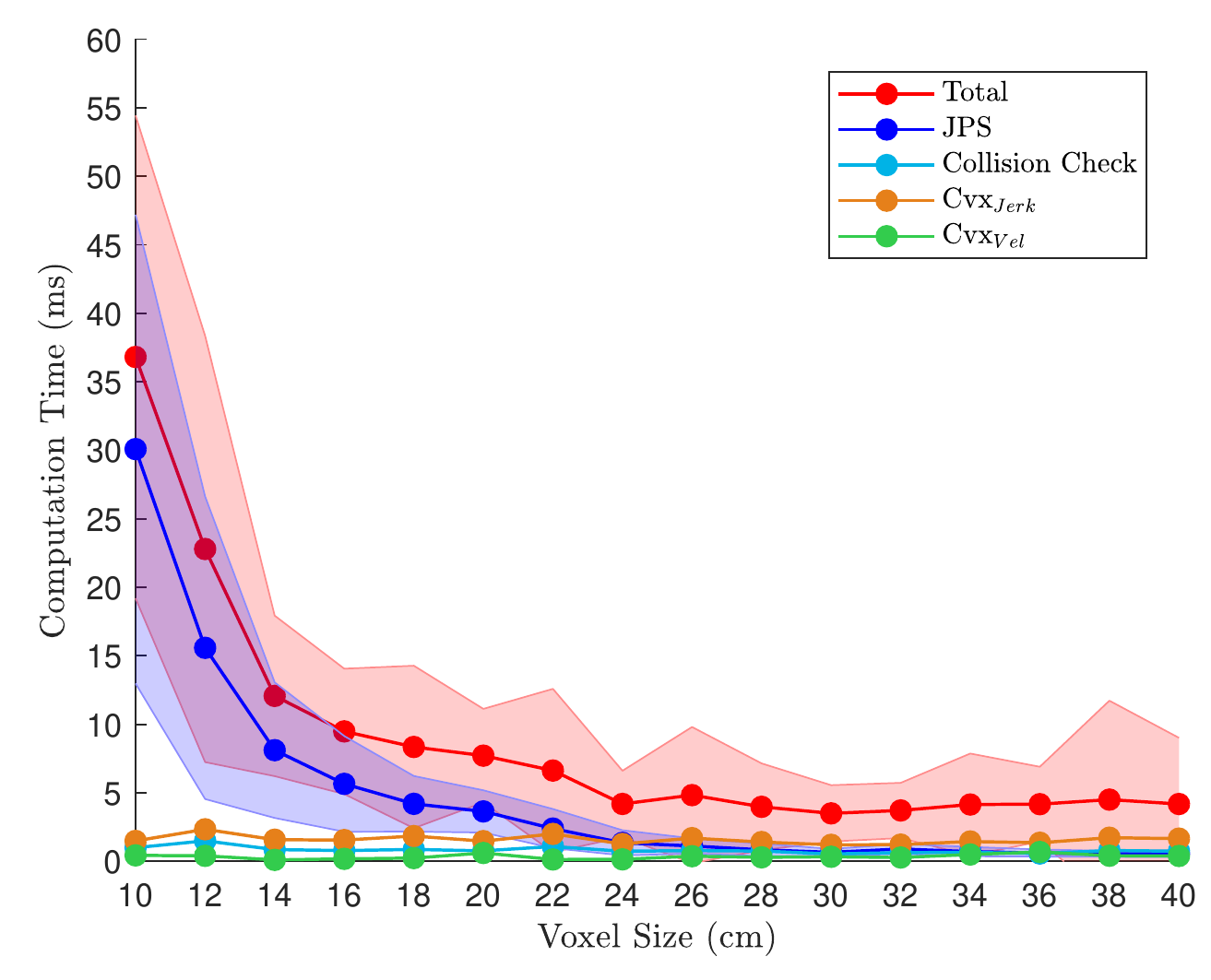}
	\caption{Timing breakdown for different voxel sizes in the forest simulation. The replan function takes less than 10 ms when the voxel size is bigger than 15 cm. The sliding map is $20$m$\times 20$m.}
	\label{fig:timing_breakdown}
\end{figure}

The map and voxel size chosen have a strong impact on the computation times and performance. For a given map, a small voxel size provides a more accurate solution, at the expense of more computation time running JPS and collision check. For a map size of $20$m$\times 20$m, the timing breakdown of the replan function in the forest simulation for different map voxel sizes is shown in Fig.~\ref{fig:timing_breakdown}. The replan function takes 37 ms in average when the voxel size is 10 cm, and is reduced to less than 10 ms when the voxel size is bigger than 15 cm. JPS takes $\approx 80\%$ of the total replanning time when the voxel size is low, and $\approx 15\%$ when the voxel size is higher. This is due to the fact that the computation time of JPS depends on the number of cells, which is reduced by the cube of the voxel size.  Note that the values of Cvx$_{jerk}$ and Cvx$_{vel}$ indicate the total computation time for all the jerk-controlled and velocity-controlled primitives computed in each replanning step. The mean time per primitive is 1.3 ms for jerk and 0.3 ms for velocity. The distances of these primitives range from 0.5 to 4 m and from 2 to 5.5 m (depending on the geometry of the obstacles) for the jerk-controlled and velocity-controlled primitives respectively.

\begin{table}[t]
	\caption{\label{tab:table_forest}Distances obtained by seven different methods in the cluttered forest simulation. The distance values are computed for the cases that reach the goal. The improvement percentages are computed for the minimum and the maximum of each column. The results for the other planners were provided by the authors of \cite{oleynikova2018safe}.}
	\begin{tabular}{p{1.5cm} >{\centering\arraybackslash}p{1.5cm} >{\raggedleft\arraybackslash}p{0.75cm} >{\raggedleft\arraybackslash}p{0.75cm} >{\raggedleft\arraybackslash}p{0.75cm} >{\raggedleft\arraybackslash}p{0.75cm}}
		\hline
		\hline
		\multicolumn{1}{l}{\textbf{Method}} & \multicolumn{1}{l}{\textbf{Number of}}     & \multicolumn{4}{c}{\textbf{Distance (m)}}                                                            \\ \cline{3-6} 
		\multicolumn{1}{l}{} & \textbf{Successes} & \textbf{Avg}  & \textbf{Std}  & \multicolumn{1}{l}{\textbf{Max}} & \multicolumn{1}{l}{\textbf{Min}} \\ 
		Incremental                         & 0                        & -             & -             & -                                & -                                 \\ 
		Rand. Goals                        & \textbf{10}              & 138.0         & 32.0          & 210.5                            & 105.6                             \\ 
		Opt. RRT$^\star$                     & 9                        & 105.3         & \textbf{10.3}          & 126.4                            & 95.5                              \\ 
		Cons. RRT$^\star$                   & 9                        & 155.8         & 52.6          & 267.9                            & 106.2                             \\ 
		NBVP  \cite{bircher2016receding}                              & 6                        & 159.3         & 45.6          & 246.9                            & 123.6                             \\ 
		SL Expl. \cite{oleynikova2018safe}                   & 8                        & 103.8         & 21.6          & 148.3                            & 86.6         \\ 
				\textbf{Ours}                  & \textbf{10}              & \textbf{84.5} & 11.7 & \textbf{109.4}                   & \textbf{73.2}       \\ \hline
	 \multicolumn{2}{l}{\hspace*{-.5em} \rule{0pt}{10pt} \textbf{Min/Max improvement (\%)}}                          & \textbf{19/47}              & \textbf{-14/78} & \textbf{13/59} & \textbf{16/41}                                       \\ \hline \hline
\end{tabular}
	
\end{table}

To compare the trajectory found by our approach (in which the map is \textbf{discovered} as the UAV proceeds) with the optimal trajectory when the map is completely \textbf{known}, we use two simulation environments: a \textit{bugtrap} scenario (Fig.~\ref{fig:upenn}) and a cluttered office scenario (Fig.~\ref{fig:office}). In the \textit{bugtrap} scenario, our method produces a trajectory of $56.8$ m, approximately the same length as the optimal trajectory ($56.3$ m). In the office simulation, the total length with our approach is $41.5$ m (optimal trajectory is $35.9$ m). In this simulation, the UAV enters two rooms, but when it detects that there is no exit, turns back and finds another path to the goal. In both simulations, the optimal trajectory has been obtained using the approach proposed in \cite{liu2018searchBasedSE3}, which is optimal in the discretized space. As in our approach the world is being discovered gradually, our solution is not globally optimal, and it requires more control effort than the optimal one. However, the similarity between these two paths reflects the performance of our algorithm, able to obtain a near-optimal path even when the world is discovered gradually.

\subsection{Hardware Experiments}

\begin{figure}[]
	\centering
	\includegraphics[width=\columnwidth]{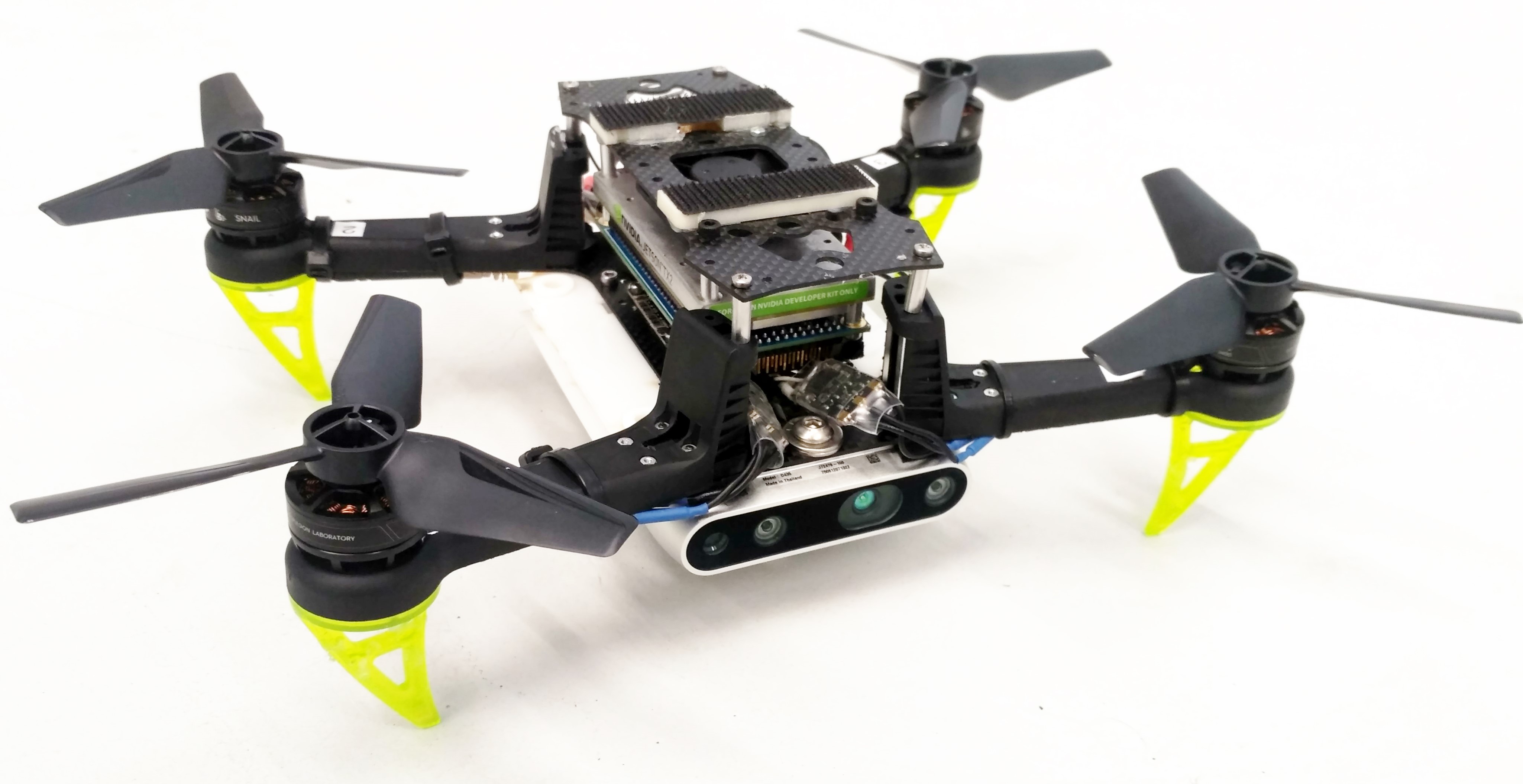}
	\caption{UAV used in the experiments. It is equipped with a Qualcomm\textsuperscript{\textregistered} SnapDragon Flight, an Nvidia\textsuperscript{\textregistered} Jetson TX2 and an Intel\textsuperscript{\textregistered} RealSense Depth Camera D435.  }
	\label{fig:drone}
\end{figure}

The UAV used for the hardware experiments is shown in Fig. \ref{fig:drone}. All the perception, planning and control runs onboard, and the position, velocity, attitude, and IMU biases are estimated by fusing propagated IMU measurements with an external motion capture system via a Kalman filter. The mapping fusion times achieved onboard are $50$ms and $80$ms for depth image resolutions of $480\times 270$ and $640\times 480$ respectively. 
 All these experiments are available in
 \url{https://www.youtube.com/watch?v=E4V2_B8x-UI}.

\section{CONCLUSIONS}
\label{sec:conclusions_future_work}
This work presented a novel planning a mapping framework suitable for agile flights in unknown environments. The key properties of this framework is its ability to solve the interaction between the global planner and the local planner considering the dynamics of the vehicle, and its ability to address efficiently the integration between a fast planner and a slower mapper. The replanning and mapping rates are several times faster than the state of the art. 

The Gazebo worlds and the code for the optimizer are available at \url{https://github.com/jtorde/}.





\section*{ACKNOWLEDGMENT}

Thanks to Boeing Research \& Technology for support of
the hardware, to Helen Oleynikova (ASL-ETH) for the data of the forest simulation, and to Pablo Tordesillas (ETSAM-UPM) for his help with some figures of this paper. Supported in part by Defense Advanced Research Projects Agency (DARPA) as part of the Fast Lightweight Autonomy (FLA) program, HR0011-15-C-0110.
Views expressed here are those of the authors, and do not reflect the official views or policies of the Dept. of Defense or the U.S. Government.

\newpage

\balance

\makeatletter
\def\endthebibliography{%
	\def\@noitemerr{\@latex@warning{Empty `thebibliography' environment}}%
	\endlist
}
\makeatother

\bibliographystyle{unsrt}
\bibliography{ref}

\end{document}